\documentclass{article} 
\usepackage{iclr2022_conference,times}

\usepackage{amsmath,amsfonts,bm}

\def\eqref#1{equation~\ref{#1}}

\def\1{\bm{1}}

\DeclareMathAlphabet{\mathsfit}{\encodingdefault}{\sfdefault}{m}{sl}
\SetMathAlphabet{\mathsfit}{bold}{\encodingdefault}{\sfdefault}{bx}{n}

\def\sS{{\mathbb{S}}}

\def\sW{{\mathbb{W}}}

\usepackage{hyperref}
\usepackage{url}

\usepackage[ruled]{algorithm2e}
\usepackage{multirow}
\usepackage{booktabs}
\usepackage{graphicx}

\newcommand{\eg}{{\em e.g.}}
\newcommand{\ie}{{\em i.e.}}

\title{Optimizing Neural Networks with\\Gradient Lexicase Selection}

\author{Li Ding\\
University of Massachusetts Amherst\\
\texttt{liding@umass.edu} \\
\And
Lee Spector\\
Amherst College\\
University of Massachusetts Amherst\\
\texttt{lspector@amherst.edu} \\
}

\iclrfinalcopy 
\begin{document}

\maketitle

\begin{abstract}
    One potential drawback of using aggregated performance measurement in machine learning is that models may learn to accept higher errors on some training cases as compromises for lower errors on others, with the lower errors actually being instances of overfitting.
    This can lead to both stagnation at local optima and poor generalization.
    Lexicase selection is an uncompromising method developed in evolutionary computation, which selects models on the basis of sequences of individual training case errors instead of using aggregated metrics such as loss and accuracy.
    In this paper, we investigate how lexicase selection, in its general form, can be integrated into the context of deep learning to enhance generalization.
    We propose Gradient Lexicase Selection, an optimization framework that combines gradient descent and lexicase selection in an evolutionary fashion.
    Our experimental results demonstrate that the proposed method improves the generalization performance of various widely-used deep neural network architectures across three image classification benchmarks.
    Additionally, qualitative analysis suggests that our method assists networks in learning more diverse representations.
    Our source code is available on GitHub:
    \url{https://github.com/ld-ing/gradient-lexicase}.
\end{abstract}

\section{Introduction}

Modern data-driven learning algorithms, in general, define an optimization
objective, \eg, a fitness function for parent selection in genetic
algorithms~\citep{holland1992genetic} or a loss function for gradient descent in
deep learning~\citep{lecun2015deep}, which computes the aggregate performance on
the training data to guide the optimization process. Taking the image
classification problem as an example, most recent approaches use Cross-Entropy
loss with gradient descent~\citep{bottou2010large} and
backpropagation~\citep{rumelhart1985learning} to train deep neural networks
(DNNs) on batches of training images. Despite the success that advanced DNNs can
reach human-level performance on the image recognition
task~\citep{russakovsky2015imagenet}, one potential drawback for such aggregated
performance measurement is that the model may learn to seek ``compromises''
during the learning procedure, \eg, optimizing model weights to intentionally
keep some errors in order to gain higher likelihood on correct predictions. To
give an example, consider a situation that may happen during the training phase
of image classification for a batch of 10 images: 9 of them are correctly
predicted with high probabilities, but one is wrong. The aggregated loss may
produce gradients that guide the model weights to compromise the wrong case for
higher probabilities on other cases, which may lead to the optimization process
getting stuck at local optima.

We refer to problems for which such compromises are undesirable as
\textit{uncompromising problems}~\citep{helmuth2014solving}, that is, as
problems for which it is not acceptable for a solution to perform sub-optimally
on any one of the cases in exchange for better performance on others. In deep
learning, in order to improve the generalization~\citep{zhang2017understanding}
of DNNs, it is important to maintain the diversity and generality of the
representations contributed by every training case.

From the literature, uncompromising problems have been recently explored in
genetic programming (GP) and genetic algorithms (GAs) for tasks such as program
synthesis. Among many methods that aim to mitigate this problem, lexicase
selection~\citep{helmuth2014solving,spector2012assessment} has been shown to
outperform many other
methods~\citep{fieldsend2015strength,galvan2013using,krawiec2015automatic} in a
number of applications and benchmarks~\citep{helmuth2015general,helmuth2021psb2}.
Instead of using an aggregated fitness function for parent selection, lexicase
selection gradually eliminates candidates as it proceeds to look at how the
population fares at each data point in the shuffled training dataset, in which
way it can bolster the diversity and generality in populations. Recent works
also show that lexicase selection can be used in rule-based learning
systems~\citep{aenugu2019lexicase}, symbolic regression~\citep{la2016epsilon},
constraint satisfaction problems~\citep{metevier2019lexicase}, machine
learning~\citep{la2020learning,la2020genetic}, and evolutionary
robotics~\citep{huizinga2018evolving,la2018behavioral} to improve model
generalization, especially in situations of diverse and unbalanced data. It is
reasonable to suspect that for many deep learning problems such as image
classification, due to natural variances in real-world data collection, lexicase
selection is likely to help improve the generalization of models.

In this work, we aim to explore the application of lexicase selection in the
task of optimizing deep neural networks. Taking advantage of the commonly-used
gradient descent and backpropagation methods, we introduce Gradient Lexicase
Selection, an optimization framework for training DNNs that not
only benefit from the efficiency of gradient-based learning but also improves
the generalization of the networks using the outline of lexicase selection
method in an evolutionary fashion. We test the proposed method on the basic
image classification task on three benchmark datasets
(CIFAR-10~\citep{krizhevsky2009learning},
CIFAR-100~\citep{krizhevsky2009learning}, and SVHN~\citep{netzer2011reading}).
Experimental results show that gradient lexicase selection manages to improve
the performance of the DNNs consistently across six different popular
architectures (VGG~\citep{simonyan15very}, ResNet~\citep{he2016deep},
DenseNet~\citep{huang2017densely}, MobileNetV2~\citep{sandler2018mobilenetv2},
SENet~\citep{hu2018squeeze}, EfficientNet~\citep{tan2019efficientnet}). In
addition, we perform further ablation studies to analyze the effectiveness and
robustness of the proposed method from various perspectives. We first introduce
variants to our method by using random selection and tournament selection, in
order to validate the contribution of each component in the framework. We also
investigate the trade-offs between exploration and exploitation by analyzing the
effects of changing population size and momentum. Finally, the qualitative
analysis shows that our algorithm produces better representation diversity,
which is advantageous to the generalization of DNNs.

\section{Background and Related Work}

\textbf{Preliminaries of Lexicase Selection}\hspace{1em}

Lexicase selection is initially proposed as a parent selection method in
population-based stochastic search algorithms such as genetic
programming~\citep{helmuth2014solving,spector2012assessment}. Follow-up work has
shown that it can effectively improve behavioral diversity and
the overall performance and on a variety of genetic programming
problems~\citep{helmuth2016lexicase,helmuth2015general,helmuth2014solving,liskowski2015comparison}.
The key idea is that each selection event considers a
randomly shuffled sequence of training cases. As a result, lexicase selection
sometimes selects specialist individuals that perform poorly on average but
perform better than many individuals on one or more other cases. We include a more detailed description of lexicase selection in Appendix~\ref{sec:lexicase}.

Unlike methods such as tournament selection that use a single fitness value and
thus tend to always select generalist individuals that have good average
performance, lexicase selection does not base selection on an aggregated measure of
performance. Such a difference allows lexicase selection to maintain higher
population diversity by prioritizing different parts of the dataset during each
selection event through the ordering of the cases. It has been shown empirically
on a number of program synthesis benchmark problems that lexicase selection
substantially outperforms standard tournament selection and typically maintains
higher levels of diversity~\citep{helmuth2016lexicase}.

In a more general context, lexicase selection can be used in any case where a
selection procedure occurs with regard to performance assessment of multiple
candidates with a set of training cases. Recent work also explores the usage of
lexicase selection in rule-based learning systems~\citep{aenugu2019lexicase},
symbolic regression~\citep{la2016epsilon}, constraint satisfaction
problems~\citep{metevier2019lexicase}, machine
learning~\citep{la2020learning,la2020genetic}, and evolutionary
robotics~\citep{huizinga2018evolving,la2018behavioral}. In this work, we aim to
explore the effectiveness of lexicase selection in the context of deep learning
optimization from the perspective of improving the diversity of gradient-based
representation learning for better generalization. While there are also other
parent selection
methods~\citep{fieldsend2015strength,galvan2013using,krawiec2015automatic} that
have been proposed to achieve similar goals, in this work we focus on
investigating the usage of lexicase selection in deep learning. Further
discussion of comparisons between lexicase selection to other selection methods
are out of the scope of this work.

\textbf{Deep Neuroevolution and Population-based Optimization}\hspace{1em}

While backpropagation~\citep{rumelhart1985learning} with gradient descent has
been the most successful method in training DNNs with fixed-topology in the past
few decades~\citep{lecun2015deep}, there are also attempts to train DNNs through
evolutionary algorithms (EAs). \citet{such2017deep} proposed a gradient-free
method to evolve the network weights by using a simple genetic algorithm, and
was able to evolve a relatively deep network (with 4 million parameters) and
demonstrated competitive results on several reinforcement learning benchmark
problems. \citet{jaderberg2017population} proposed population-based training to
optimize both model and the hyperparameters. \citet{cui2018evolutionary}
proposed to alternate between the SGD step and evolution step to improve the
average fitness of the population. \citet{ding2021evolving} demonstrate the
usage of selection-inspired methods as regularization of DNNs. The most relevant
recent work is \citet{pawelczyk2018genetically}, which did a pilot study on
combining simple GA schema with gradient-based learning, where gradient training
is used as part of the mutation process. Although the method was tested only on
one dataset, the results were encouraging and offered some insights on a deeper
combination of GAs and gradient learning. Following this trend, this paper aims
to explore a more efficient evolutionary framework that takes advantage of both
SGD and lexicase selection, to improve the network generalization by treating
image recognition as an uncompromising problem.

From a broader perspective that is also closely related to this work, there has
been a surge of interest in methods for Neural Architecture Search
(NAS)~\citep{elsken2019neural}, where evolutionary algorithms gain high
popularity. A majority of
methods~\citep{floreano2008neuroevolution,liu2017hierarchical,miikkulainen2019evolving,real2019regularized,real2017large,stanley2002evolving,xie2017genetic}
use EA to search neural network topologies and use backpropagation to optimize
network weights, and some others~\citep{stanley2002evolving} use EA to co-evolve
topologies along with weights. While our method can be easily extended to the
NAS problem, this work focuses on training various fixed-topology networks in
order to make fair comparisons to show the significance of using lexicase
selection to improve model generalization.

\section{Gradient Lexicase Selection}

\begin{figure}[t]
    \begin{center}
        \includegraphics[width=\textwidth]{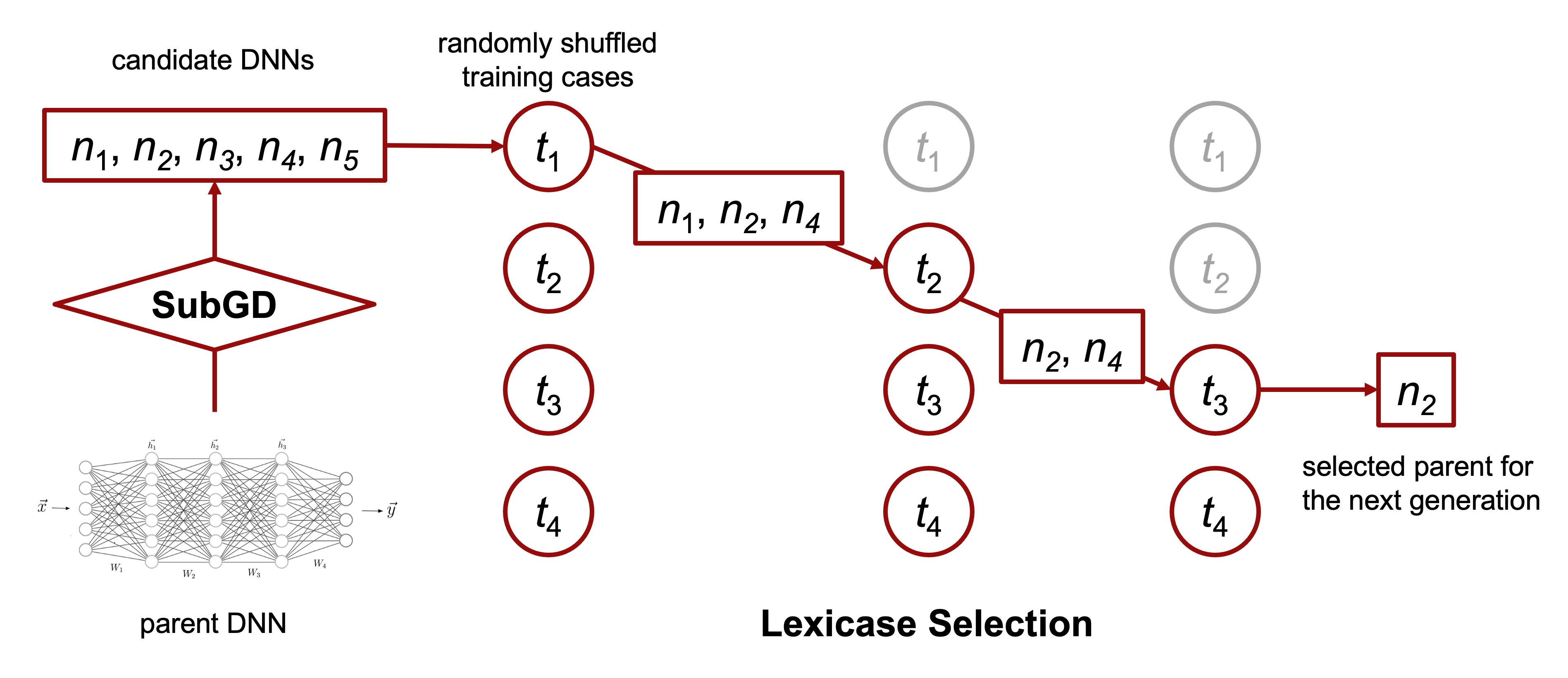}
    \end{center}
    \caption{Overview of the proposed gradient lexicase selection. Given the parent model, we first generate candidates by running subset gradient descent (SubGD), then perform lexicase selection by assessing candidates on each individual case to obtain the parent model for the next generation.}
    \label{fig:alg}
\end{figure}

Our goal is to integrate lexicase selection to improve the generalization of
DNNs, while at the same time to the greatest extent keep the efficiency of the
popular gradient-based learning. We propose Gradient Lexicase Selection as an
optimization framework to combine the strength of these two methods. The
algorithm is outlined in Alg.~\ref{alg:gradlexi}. An overview of the algorithm
is also depicted in Fig.~\ref{fig:alg}. The proposed algorithm has two main
components, Subset Gradient Descent (SubGD) and Lexicase Selection, which we
describe in details in this section as follows.

\subsection{Evolution with Subset Gradient Descent (SubGD)}

First, we introduce the general evolutionary framework that uses a combination
of stochastic gradient descent and evolution. Given the network topology, we
first initialize all the parameters $\sW_0$ as the initial parent weights. For
each generation, given a population size of $p$, we generate $p$ instances of
the model as $p$ offspring with the same weights as the parent weights, namely,
$\sW^{(0)}=\sW^{(1)}=\cdots=\sW^{(p-1)}=\sW_0$. We then perform mutation on
these offspring and use lexicase selection to select the parent for the next
generation.

For each generation, we have $p$ instances of the model starting with the exact
same weights. Instead of random mutations such as adding gaussian noise as
commonly used in neuroevolution, we propose a gradient-based mutation method
called subset gradient descent (SubGD).

Given the whole training dataset $\sS_{train}$, we divided it into $p$ subsets
with random sampling, as $\sS_{train}^{(0)}, \sS_{train}^{(1)}, \cdots,
    \sS_{train}^{(p-1)}$. We then train each model instance accordingly on one of
the subsets using the normal mini-batch stochastic gradient descent. The
mutation is done when all the training data is consumed, which is one epoch in
traditional deep learning.

There are several advantages of the proposed mutation method. First, since all
the offspring are trained with different non-overlapping training samples, they
are likely to evolve diversely, especially when data augmentation is also
included. Secondly, each off-spring is trained using gradient descent, meaning
they will be optimized efficiently towards the objective, comparing to random
mutation methods such as gaussian noise. Thirdly, if implemented with
distributed training, all the offspring can be trained simultaneously to further
reduce computation time. In general, the subset gradient descent aims to find a
balance between exploration and exploitation during the evolution process for
more efficient optimization.

\begin{figure}[t]
    \centering
    \begin{minipage}{.9\linewidth}
        \begin{algorithm}[H]
            \KwData{
                \begin{itemize}
                    \item \texttt{data} - the whole training dataset
                    \item \texttt{candidates} - set of $p$ instances of the DNN model initialized with the same parameters
                \end{itemize}
            }
            \KwResult{
                \begin{itemize}
                    \item an optimized DNN model
                \end{itemize}
            }

            \tcp{K training epochs}

            \For{$epoch = 1:K$}{
                \texttt{subsets} $\gets$ $p$ equal-size subsets obtained through random sampling from the entire \texttt{data} without replacement

                Use gradient descent and backpropagation to optimize each of the $p$ \texttt{candidates} on each of the $p$ \texttt{subsets} respectively

                \texttt{cases} $\gets$ randomly shuffled sequence of \texttt{data} to be used in lexicase selection

                \texttt{parent} $\gets$ \textit{None}

                \For{\texttt{case} in \texttt{cases}}{
                    Evaluate all the \texttt{candidates} on \texttt{case}.

                    \texttt{candidates} $\gets$ the subset of the current \texttt{candidates} that have exactly best performance on \texttt{case}

                    \If{\texttt{candidates} contains only one single \texttt{candidate}}{
                        \texttt{parent} $\gets$ \texttt{candidate}

                        break
                    }
                }

                \If{\texttt{parent} is \textit{None}}{
                    \texttt{parent} $\gets$ a randomly selected individual in \texttt{candidates}
                }

                \texttt{candidates} $\gets$ set of $p$ instances of the DNN model copied with the same parameters as \texttt{parent}

            }
            \Return{\texttt{parent}}
            \caption{Gradient Lexicase Selection}
            \label{alg:gradlexi}
        \end{algorithm}
    \end{minipage}
\end{figure}

\subsection{Lexicase Selection for DNNs}

After mutation, the offspring become candidates and we use lexicase selection to
select a parent from them for the next generation. First, a randomly shuffled
sequence of training data points (without data augmentation) is used for
selection. Starting from the first training sample, we evaluate all the
candidates on each case individually and remove the candidate from the selection
pool if it does not make the correct prediction. This process is repeated until
if 1) there is only one candidate left, which will be selected as the parent for
the next generation, or 2) all the training samples are exhausted and more than
one candidates survive, in which case we randomly pick a candidate from the
selection pool.

For the selection process, we do not hold out another validation set because 1)
if we choose to use a validation set, the validation set should have an adequate
size in order to ensure its diversity and generality, which means the training
set will be noticeably smaller, and thus the training performance is likely to
degrade; 2) since each model instance only gets access to part of the
(augmented) training data, the selection performed on the original training data
is still effective, since the exact same data was never used in the mutation
(training).

An important feature of lexicase selection is that it treats all the cases
equally and thus there is no way to modify its selection pressure. The
motivation behind lexicase selection is to allow the survival of those models
which may not perform best overall but were able to solve the given testing
cases. Such a model is likely to learn essential feature representations that
allow it to make correct predictions on specific cases where all others fail. By
letting lexicase selection guide the training process, the neural network can
potentially learn more diverse representations that finally contribute to better
generalization.

To better accommodate the situation of training deep models on large-scale
datasets, we also make some slight modifications to the original lexicase
selection algorithm in regard to the tie situations, \ie, when all the remaining
candidates fail to make the correct prediction on one case. The original
lexicase selection lets all the candidates survive because they all have the
same ``best'' performance, which is failure. However, in the early stages of DNN
training, while all the candidates are unable to predict correctly on any case,
the original lexicase selection will proceed to evaluate them until someone
happens to get a correct prediction by chance, which is inefficient especially
on large datasets. So we modify the algorithm to randomly select a candidate
from the remaining candidates if they all fail. The modification improves the
efficiency of early-stage training and has not shown any noticeable effect on
the final model performance.

\section{Experimental Results}

The proposed Gradient Lexicase Selection is tested on the task of image
classification, which is one of the most common benchmark problems in computer
vision and deep learning in general. We implement the algorithm on six popular DNN
architectures (VGG~\citep{simonyan15very}, ResNet~\citep{he2016deep},
DenseNet~\citep{huang2017densely}, MobileNetV2~\citep{sandler2018mobilenetv2},
SENet~\citep{hu2018squeeze}, EfficientNet~\citep{tan2019efficientnet}). To show
the significance of our method, we also implement the original momentum-SGD
training as baselines for all the architectures.

Three benchmark datasets (CIFAR-10~\citep{krizhevsky2009learning},
CIFAR-100~\citep{krizhevsky2009learning}, and SVHN~\citep{netzer2011reading}) are
used for evaluation. These datasets comprise $32 \times 32$ pixel real-world RGB
images of common objects (CIFAR-10, CIFAR-100) and street scene digits (SVHN).
The training is done using the training set only and we evaluate the methods on
the test set after the training process is finished. Note that for illustration
purposes, we only use the training dataset of SVHN without the large
\textit{extra} set, so the results are not comparable to other work.

\subsection{Image Classification Results}\label{sec:cls}

The image classification results are shown in Tab.~\ref{tab:cls}. We report the
mean percentage accuracy (\textit{acc.}) with standard deviation (\textit{std.})
obtained by running the same experiment with three different random seeds. The
last column (\textit{acc.} $\uparrow$) calculates the difference of accuracy by
using our method compared to baseline, where positive numbers indicate
improvement. We can first see that by using our method, most of the
architectures show significant improvement on the testing result. On the easier
SVHN dataset, we can still observe moderate and consistent improvement. To show
the robustness of our algorithm, we use the same population size of $4$ for
lexicase in all the experiments, meaning the performance may be further improved
if extra tuning is performed. The ablation study on the effect of population
size is described later in Sec.\ref{sec:pop}.

\begin{table}[t]
    \caption{Image classification results. We report the mean percentage accuracy (\textit{acc.}) with standard deviation (\textit{std.}) obtained by running the same experiment with three different random seeds. The last column (\textit{acc.} $\uparrow$) calculates the difference of accuracy by using our method compared to baseline, where positive numbers indicate  improvement. }
    \label{tab:cls}
    \centering
    \begin{tabular}{llccccc}
        \toprule
        \multirow{2}{*}{Dataset}   & \multirow{2}{*}{Architecture} & \multicolumn{2}{c}{Baseline} & \multicolumn{2}{c}{Lexicase} & \multicolumn{1}{l}{}                                            \\
        \cmidrule(r){3-4}
        \cmidrule(r){5-6}
                                   &                               & \textit{acc.}                & \textit{std.}                & \textit{acc.}        & \textit{std.} & \textit{acc.} $\uparrow$ \\
        \midrule
        \multirow{7}{*}{CIFAR-10}  & VGG16                         & 92.85                        & 0.10                         & 93.40                & 0.13          & \textbf{0.55}            \\
                                   & ResNet18                      & 94.82                        & 0.10                         & 95.35                & 0.06          & \textbf{0.53}            \\
                                   & ResNet50                      & 94.63                        & 0.46                         & 94.98                & 0.18          & \textbf{0.34}            \\
                                   & DenseNet121                   & 95.06                        & 0.31                         & 95.38                & 0.04          & \textbf{0.32}            \\
                                   & MobileNetV2                   & 94.37                        & 0.19                         & 93.97                & 0.12          & -0.39                    \\
                                   & SENet18                       & 94.69                        & 0.14                         & 95.37                & 0.23          & \textbf{0.68}            \\
                                   & EfficientNetB0                & 92.60                        & 0.18                         & 93.00                & 0.22          & \textbf{0.40}            \\

        \midrule

        \multirow{7}{*}{CIFAR-100} & VGG16                         & 72.09                        & 0.52                         & 72.53                & 0.20          & \textbf{0.44}            \\
                                   & ResNet18                      & 76.33                        & 0.29                         & 76.68                & 0.40          & \textbf{0.35}            \\
                                   & ResNet50                      & 76.82                        & 0.96                         & 77.44                & 0.25          & \textbf{0.63}            \\
                                   & DenseNet121                   & 78.72                        & 0.82                         & 79.08                & 0.26          & \textbf{0.36}            \\
                                   & MobileNetV2                   & 75.87                        & 0.28                         & 75.57                & 0.30          & -0.30                    \\
                                   & SENet18                       & 76.97                        & 0.06                         & 77.22                & 0.29          & \textbf{0.25}            \\
                                   & EfficientNetB0                & 71.03                        & 0.86                         & 71.36                & 0.87          & \textbf{0.33}            \\

        \midrule

        \multirow{7}{*}{SVHN}      & VGG16                         & 96.27                        & 0.06                         & 96.29                & 0.08          & \textbf{0.02}            \\
                                   & ResNet18                      & 96.43                        & 0.14                         & 96.62                & 0.08          & \textbf{0.19}            \\
                                   & ResNet50                      & 96.69                        & 0.21                         & 96.74                & 0.07          & \textbf{0.04}            \\
                                   & DenseNet121                   & 96.82                        & 0.16                         & 96.87                & 0.03          & \textbf{0.05}            \\
                                   & MobileNetV2                   & 96.23                        & 0.13                         & 96.26                & 0.07          & \textbf{0.03}            \\
                                   & SENet18                       & 96.62                        & 0.19                         & 96.59                & 0.11          & -0.03                    \\
                                   & EfficientNetB0                & 96.14                        & 0.12                         & 95.94                & 0.10          & -0.20                    \\

        \bottomrule
    \end{tabular}
\end{table}

Beyond those improvements, we also find that among all the architectures, our
method surprisingly fails to improve MobileNetV2 on both CIFAR-10 and CIFAR-100.
The main difference between MobileNetV2 and other architectures is that it is a
highly optimized architecture with over an order of magnitude less parameters
compared to other architectures. \citet{sandler2018mobilenetv2} stated that they
tailor the architecture to different performance points, which can be adjusted
depending on desired accuracy/performance trade-of. As a result, it is likely
that the accuracy is restricted by the model size, and even with better training
strategies the performance is not going to improve. Such results indicate that
our method may not work directly with architectures that have been optimized by
using other training methods. But for other more general architectures our
method work directly out-of-the-box without further tuning.

\subsection{Comparing Different Selection Methods} \label{sec:cls2}

The proposed method has two major components, SubGD and Lexicase Selection. To
further validate the contribution of each component, we introduce two other
selection methods for comparison: random selection and tournament
selection~\citep{miller1995genetic}. Using the same evolutionary framework with
SubGD, random selection simply selects a random offspring for each generation,
and tournament selection uses the average accuracy as the fitness function for
selection, which is an aggregated metric as oppose to lexicase selection. The
results are shown in Tab.~\ref{tab:cls2}.

\begin{table}[t]
    \caption{Comparing gradient lexicase selection to other selection methods on CIFAR-10. We report the mean percentage accuracy (\textit{acc.}) with standard deviation (\textit{std.}) obtained by running the same experiment with three different random seeds.}
    \label{tab:cls2}
    \centering
    \begin{tabular}{lcccccccc}
        \toprule
        \multirow{2}{*}{Architecture} & \multicolumn{2}{c}{SGD} & \multicolumn{2}{c}{Random} & \multicolumn{2}{c}{Tournament} & \multicolumn{2}{c}{Lexicase}                                                                  \\
        \cmidrule(r){2-3}
        \cmidrule(r){4-5}
        \cmidrule(r){6-7}
        \cmidrule(r){8-9}
                                      & \textit{acc.}           & \textit{std.}              & \textit{acc.}                  & \textit{std.}                & \textit{acc.} & \textit{std.} & \textit{acc.}  & \textit{std.} \\
        \midrule
        VGG16                         & 92.85                   & 0.10                       & 92.97                          & 0.15                         & 93.12         & 0.12          & \textbf{93.40} & 0.13          \\
        ResNet18                      & 94.82                   & 0.10                       & 94.99                          & 0.12                         & 94.90         & 0.14          & \textbf{95.35} & 0.06          \\
        ResNet50                      & 94.63                   & 0.46                       & 94.75                          & 0.13                         & 94.77         & 0.04          & \textbf{94.98} & 0.18          \\
        DenseNet121                   & 95.06                   & 0.31                       & 95.13                          & 0.04                         & 95.12         & 0.02          & \textbf{95.38} & 0.04          \\
        MobileNetV2                   & \textbf{94.37}          & 0.19                       & 94.02                          & 0.14                         & 93.91         & 0.09          & 93.97          & 0.12          \\
        SENet18                       & 94.69                   & 0.14                       & 95.04                          & 0.15                         & 95.01         & 0.23          & \textbf{95.37} & 0.23          \\
        EfficientNetB0                & 92.60                   & 0.18                       & 92.77                          & 0.11                         & 92.83         & 0.12          & \textbf{93.00} & 0.22          \\
        \bottomrule
    \end{tabular}
\end{table}

We can observe that both random selection and tournament selection perform
slightly better than the SGD baseline in most cases, but the proposed gradient
lexicase selection is consistently better than both methods with a significant
margin. Random selection can be viewed as a baseline method that uses SGD with
the same amount of computation as gradient lexicase selection, which indicates
that the proposed method outperforms SGD even at the same level of computation.
Tournament selection is one of the most commonly used selection method in
evolutionary algorithms, which select parents based on an aggregated fitness
evaluation. As the performance of tournament selection is similar to random
selection, indicating that the mechanism of lexicase selection has the major
contribution to the improvement.

\section{Ablation Studies}

In this section, we design several ablation studies to further analyze and validate the effectiveness of the proposed method. Unless specifically mentioned, all the implementation details follow the same practices in Sec.~\ref{sec:imp}.

\subsection{Population Size} \label{sec:pop}

Population size is an essential hyperparameter for evolutionary algorithms. In
this work, the population size of DNNs has more constraints such as computation
cost and total GPU memory, so it has to be much smaller comparing to those for
classic GP problems. We test lexicase gradient selection with population sizes
of $2,4,6,8$. For illustration purposes, two architectures (VGG16 and ResNet18)
are evaluated on the CIFAR-10 dataset. The results are shown in
Tab.~\ref{tab:pop}.

First, we can see that lexicase is relatively robust to different population
size $p$. Under all the population size configurations lexicase manages to
outperform baseline significantly. Since there have not been a trend of
increased accuracy with larger population size, the generalization performance
does not seem to increase with a larger population. This observation aligns well
with the behavior of lexicase selection in GP
problems~\citep{helmuth2018program,la2019probabilistic}, where there seems to be
an optimal population size through the trade-off between exploration and
exploitation.

For our method, having a larger population size not only adds more offspring for
each generation, but also reduces the size of each subset used for training each
offspring by SubGD. In either way, the exploration is reduced and the
exploitation is increased. There is no conflict between the two effects, so we
do not control the size of each subset when increasing the population size. If
the population gets too large, the individuals may not evolve different-enough
behaviors from each other, and thus the diversity of population may become
lower. In general, we find that with relatively small population size, we can
get good results for gradient lexicase selection.

\begin{table}[t]
    \caption{Comparing different population sizes on CIFAR-10. Lexicase is relatively robust to different population size $p$, and it manages to outperform baseline with all the configurations in this experiment. The generalization performance does not seem to increase with larger population size. }
    \label{tab:pop}
    \centering
    \begin{tabular}{lccccc}
        \toprule
        \multirow{2}{*}{Architecture} &          & \multicolumn{4}{c}{Lexicase with population size $p$}                                  \\
        \cmidrule(r){3-6}
                                      & Baseline & $p=2$                                                 & $p=4$ & $p=6$          & $p=8$ \\
        \midrule
        VGG16                         & 92.85    & 93.61                                                 & 93.40 & \textbf{93.92} & 93.37 \\
        ResNet18                      & 94.82    & \textbf{95.50}                                        & 95.35 & 95.27          & 95.38 \\

        \bottomrule
    \end{tabular}
\end{table}

\subsection{Momentum} \label{sec:mom}

In deep learning, SGD with
momentum~\citep{sutskever2013importance,liu2020improved} has been one of the
most widely adopted methods for training DNNs. Momentum accelerates gradient
descent by accumulating a velocity vector in directions of persistent reduction
in the objective across iterations. This accumulating behavior actually
interferes with the proposed gradient lexicase selection algorithm, because
model instances in the population get different gradient updates, and thus will
have different momentum parameters.

To solve this issue, we propose three options: 1) No Momentum: do not use
momentum at all; 2) Reset Momentum: use high momentum rate and re-initialize the
momentum parameters every epoch for each model instance; 3) Inherit Momentum:
when selecting the parent model, also copy the momentum parameters along with
the model parameters to all the instances in the next generation. For this
study, we also test two architectures (VGG16 and ResNet18) with population size
of $4$ on the CIFAR-10 dataset. The results are shown in Tab.~\ref{tab:mom}.

The key idea of lexicase selection is to select parent by using a sequence of
training cases that are prioritized lexicographically for each generation. In
this way, the population can maintain a high level of diversity. On the other
hand, momentum tends to find an aggregated direction of gradient update
accumulated through time. From the results, we can see that the Reset Momentum
option works the best, indicating that if we inherit momentum, it will influence
the mutation over generations and thus the selection strength of lexicase will
be negatively affected. By resetting momentum each epoch, only the mutation in
the current generation is accelerated by using momentum SGD, which results in a
higher diversity of offspring. In general, the momentum options can also be
viewed as different trade-offs of exploration and exploitation.

\begin{table}[t]
    \caption{Comparing different momentum configurations on CIFAR-10. Resetting
        momentum after each selection event avoids too much aggregation of
        gradients, which results in a higher diversity of offspring and thus better
        generalization performance.}
    \label{tab:mom}
    \centering
    \begin{tabular}{lcccc}
        \toprule
        \multirow{2}{*}{Architecture} &          & \multicolumn{3}{c}{Lexicase with different momentum options}                                     \\
        \cmidrule(r){3-5}
                                      & Baseline & No Momentum                                                  & Reset Momentum & Inherit Momentum \\
        \midrule
        VGG16                         & 92.85    & 92.95                                                        & \textbf{93.40} & 93.13            \\
        ResNet18                      & 94.82    & 94.77                                                        & \textbf{95.35} & 95.23            \\

        \bottomrule
    \end{tabular}
\end{table}

\subsection{Representation Diversity}

\begin{figure}[h]
    \begin{center}
        \includegraphics[width=.49\textwidth]{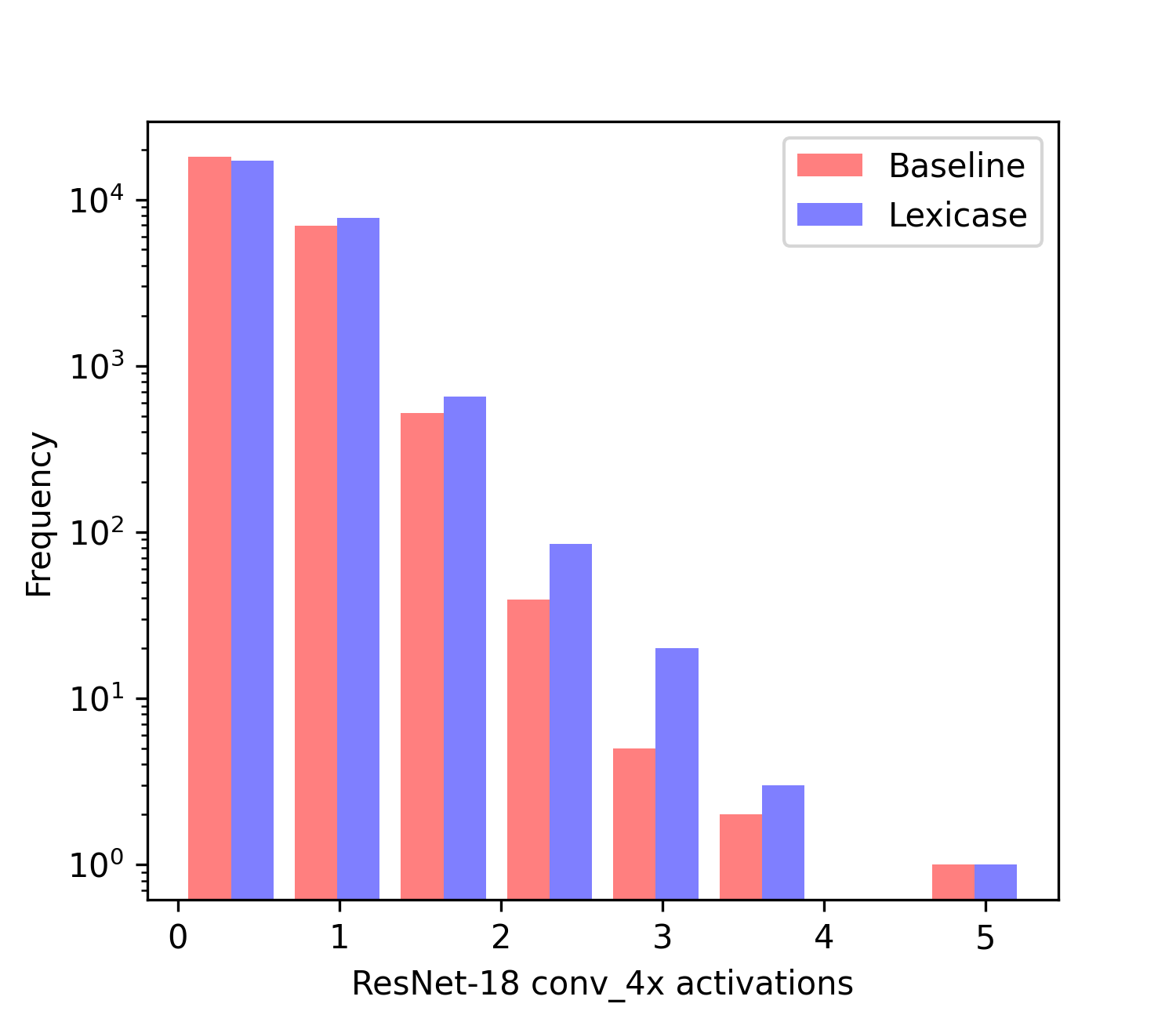}
        \includegraphics[width=.49\textwidth]{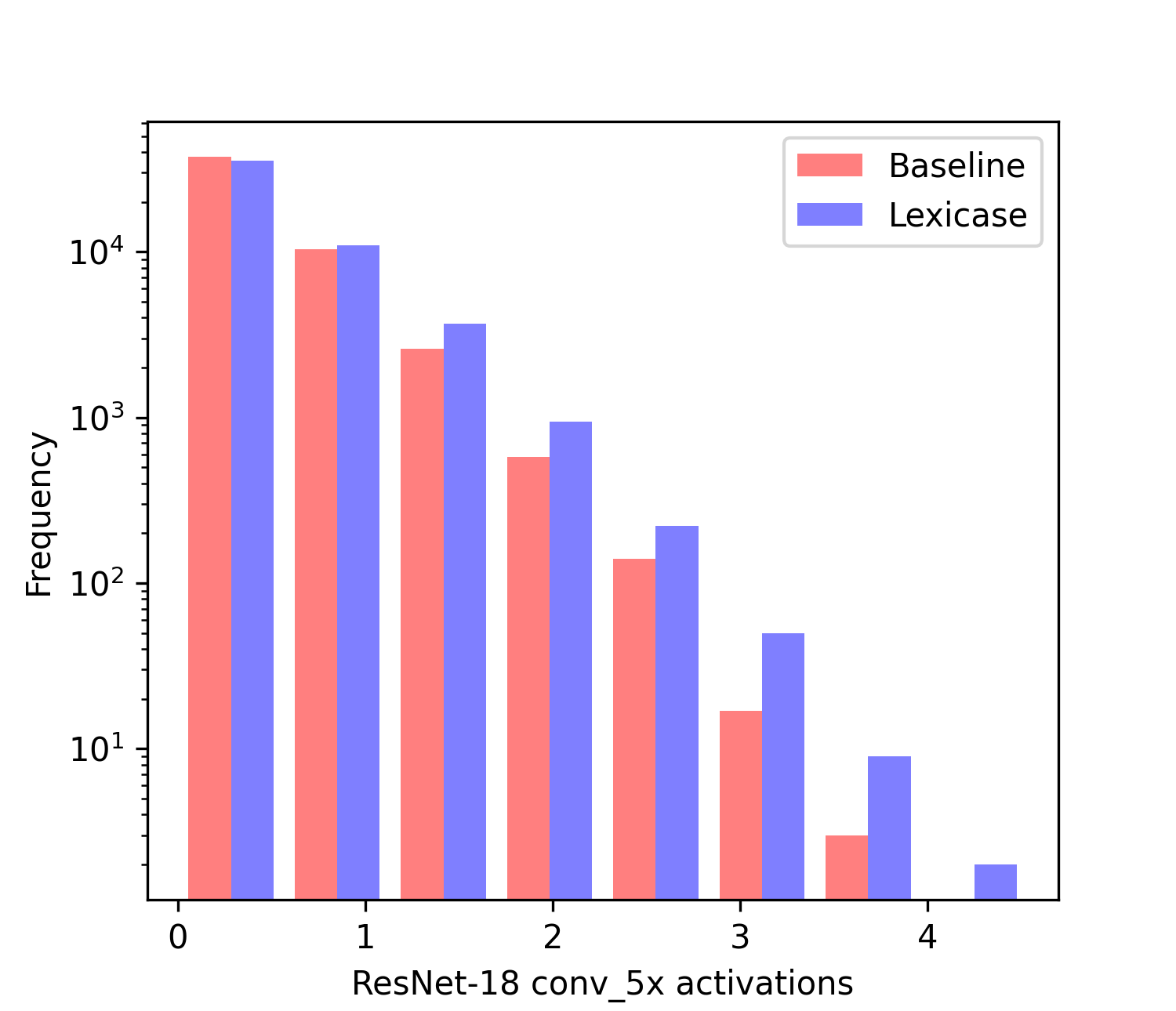}
    \end{center}
    \caption{Comparing representation diversity of normal SGD (Baseline in red) and gradient lexicase selection (lexicase in blue). The flatter distribution shows that our method produces more diverse representations.}
    \label{fig:act}
\end{figure}

While quantitative results have shown that the proposed method manages to improve
the generalization of DNNs, we would like to further investigate the reasons
behind this in order to better understand the behavior of the algorithm. One
hypothesis is that since lexicase selection is able to increase the diversity
and generality of the population in GP, it may as well help DNNs learn more
diverse representations, which improves the overall model generalization. To
validate this, we analyze and compare the feature representations in ResNet-18
trained using normal SGD and gradient lexicase selection. We take the first 100
samples from the CIFAR-10 test set and use global max pooling to obtain the
channel-wise activations of \texttt{conv\_4x} and \texttt{conv\_5x} layers (as
defined in \citet{he2016deep}). Fig.~\ref{fig:act} shows the results.

We can observe that our method produces a flatter distribution of activations
with less frequency on $0$s and more frequency on other values.
\citet{ioffe2015batch,wu2018group} shows that normalized distribution of layer
activations can help reduce the internal covariate shift of DNNs during
training, and thus improves the training efficiency and model generalization.
Similarly, our method manages to learn more diverse representations by
incorporating lexicase selection into the training framework, which is advantageous
to the generalization of DNNs.

\section{Conclusion and Future Work}

In this work, efficient adaption of lexicase selection in the task of optimizing
deep neural networks is explored. We propose gradient lexicase selection, an
evolutionary algorithm that incorporates lexicase selection with stochastic
gradient descent to help DNNs learn more diverse representations for better
generalization. Experimental results show that the proposed method can improve
the performance of several popular DNN architectures on benchmark image
classification datasets. Several ablation studies further validate the
robustness and advantages of our method from different perspectives. More
specifically, we investigate the trade-offs between exploration and exploitation
by analyzing the effects of population size and momentum. We also show that our
algorithm can produce better representation diversity, which is advantageous to
the generalization of DNNs.

The goal of our method is to improve the generalization performance rather than
speed up the optimization. Our method is potentially valuable to many real-world
problems, especially those safety-critical applications like autonomous
vehicles, where higher cost of computation during training is acceptable for
better generalization performance. There are also several factors to consider
regarding the computation cost: 1) our method expects parallel training of model
instances, so the optimal training time can be reduced to naive SGD training
with modern cloud computing facilities; 2) we have shown that with relatively
small population sizes ($4\times$ naive SGD), our method can already achieve
significantly better performance; 3) with the same amount of computation, the
naive method (random selection baseline in Sec.\ref{sec:cls2}) can not achieve
the same performance as ours.

There are several limitations of our work. As described in Sec.~\ref{sec:cls},
the current gradient lexicase selection method may not work with architectures
that have been extensively optimized, indicating a potential correlation between
network architecture and lexicase selection. For future work, we hope to explore how lexicase selection can be used in optimizing
neural architectures along with their parameters, and the integration of lexicase selection in neural architecture search in general.

\subsection*{Acknowledgments}

This material is based upon work supported by the National Science Foundation
under Grant No. 1617087. Any opinions, findings, and conclusions or
recommendations expressed in this publication are those of the authors and do
not necessarily reflect the views of the National Science Foundation.
This work was performed in part using high performance computing equipment
obtained under a grant from the Collaborative R\&D Fund managed by the
Massachusetts Technology Collaborative.
The authors would like to thank Ryan Boldi, Edward Pantridge, Thomas Helmuth,
and Anil Saini for their valuable comments and helpful suggestions.

\subsection*{Reproducibility Statement}

We submit our source code as the supplementary material for the review process,
which can be used to reproduce the experimental results in this work. We also
release our source code on Github:
\url{https://github.com/ld-ing/gradient-lexicase}. Experiment configurations and
implementation details are described in Sec.~\ref{sec:imp}.

\bibliography{iclr2022_conference}

\begin{thebibliography}{53}
\providecommand{\natexlab}[1]{#1}
\providecommand{\url}[1]{\texttt{#1}}
\expandafter\ifx\csname urlstyle\endcsname\relax
  \providecommand{\doi}[1]{doi: #1}\else
  \providecommand{\doi}{doi: \begingroup \urlstyle{rm}\Url}\fi

\bibitem[Aenugu \& Spector(2019)Aenugu and Spector]{aenugu2019lexicase}
Sneha Aenugu and Lee Spector.
\newblock Lexicase selection in learning classifier systems.
\newblock In \emph{Proceedings of the Genetic and Evolutionary Computation
  Conference}, pp.\  356--364, 2019.

\bibitem[Bottou(2010)]{bottou2010large}
L{\'e}on Bottou.
\newblock Large-scale machine learning with stochastic gradient descent.
\newblock In \emph{Proceedings of COMPSTAT'2010}, pp.\  177--186. Springer,
  2010.

\bibitem[Cui et~al.(2018)Cui, Zhang, T{\"u}ske, and
  Picheny]{cui2018evolutionary}
Xiaodong Cui, Wei Zhang, Zolt{\'a}n T{\"u}ske, and Michael Picheny.
\newblock Evolutionary stochastic gradient descent for optimization of deep
  neural networks.
\newblock \emph{Advances in Neural Information Processing Systems}, 31, 2018.

\bibitem[Ding \& Spector(2021)Ding and Spector]{ding2021evolving}
Li~Ding and Lee Spector.
\newblock Evolving neural selection with adaptive regularization.
\newblock In \emph{Proceedings of the Genetic and Evolutionary Computation
  Conference Companion}, pp.\  1717--1725, 2021.

\bibitem[Elsken et~al.(2019)Elsken, Metzen, and Hutter]{elsken2019neural}
Thomas Elsken, Jan~Hendrik Metzen, and Frank Hutter.
\newblock Neural architecture search: A survey.
\newblock \emph{The Journal of Machine Learning Research}, 20\penalty0
  (1):\penalty0 1997--2017, 2019.

\bibitem[Fieldsend \& Moraglio(2015)Fieldsend and
  Moraglio]{fieldsend2015strength}
Jonathan~E Fieldsend and Alberto Moraglio.
\newblock Strength through diversity: Disaggregation and multi-objectivisation
  approaches for genetic programming.
\newblock In \emph{Proceedings of the 2015 Annual Conference on Genetic and
  Evolutionary Computation}, pp.\  1031--1038, 2015.

\bibitem[Floreano et~al.(2008)Floreano, D{\"u}rr, and
  Mattiussi]{floreano2008neuroevolution}
Dario Floreano, Peter D{\"u}rr, and Claudio Mattiussi.
\newblock Neuroevolution: from architectures to learning.
\newblock \emph{Evolutionary intelligence}, 1\penalty0 (1):\penalty0 47--62,
  2008.

\bibitem[Galvan-Lopez et~al.(2013)Galvan-Lopez, Cody-Kenny, Trujillo, and
  Kattan]{galvan2013using}
Edgar Galvan-Lopez, Brendan Cody-Kenny, Leonardo Trujillo, and Ahmed Kattan.
\newblock Using semantics in the selection mechanism in genetic programming: a
  simple method for promoting semantic diversity.
\newblock In \emph{2013 IEEE Congress on Evolutionary Computation}, pp.\
  2972--2979. IEEE, 2013.

\bibitem[He et~al.(2016)He, Zhang, Ren, and Sun]{he2016deep}
Kaiming He, Xiangyu Zhang, Shaoqing Ren, and Jian Sun.
\newblock Deep residual learning for image recognition.
\newblock In \emph{Proceedings of the IEEE conference on computer vision and
  pattern recognition}, pp.\  770--778, 2016.

\bibitem[Helmuth \& Kelly(2021)Helmuth and Kelly]{helmuth2021psb2}
Thomas Helmuth and Peter Kelly.
\newblock Psb2: the second program synthesis benchmark suite.
\newblock In \emph{Proceedings of the Genetic and Evolutionary Computation
  Conference}, pp.\  785--794, 2021.

\bibitem[Helmuth \& Spector(2015)Helmuth and Spector]{helmuth2015general}
Thomas Helmuth and Lee Spector.
\newblock General program synthesis benchmark suite.
\newblock In \emph{Proceedings of the 2015 Annual Conference on Genetic and
  Evolutionary Computation}, pp.\  1039--1046, 2015.

\bibitem[Helmuth et~al.(2014)Helmuth, Spector, and
  Matheson]{helmuth2014solving}
Thomas Helmuth, Lee Spector, and James Matheson.
\newblock Solving uncompromising problems with lexicase selection.
\newblock \emph{IEEE Transactions on Evolutionary Computation}, 19\penalty0
  (5):\penalty0 630--643, 2014.

\bibitem[Helmuth et~al.(2016)Helmuth, McPhee, and Spector]{helmuth2016lexicase}
Thomas Helmuth, Nicholas~Freitag McPhee, and Lee Spector.
\newblock Lexicase selection for program synthesis: a diversity analysis.
\newblock In \emph{Genetic Programming Theory and Practice XIII}, pp.\
  151--167. Springer, 2016.

\bibitem[Helmuth et~al.(2018)Helmuth, McPhee, and Spector]{helmuth2018program}
Thomas Helmuth, Nicholas~Freitag McPhee, and Lee Spector.
\newblock Program synthesis using uniform mutation by addition and deletion.
\newblock In \emph{Proceedings of the Genetic and Evolutionary Computation
  Conference}, pp.\  1127--1134, 2018.

\bibitem[Holland(1992)]{holland1992genetic}
John~H Holland.
\newblock Genetic algorithms.
\newblock \emph{Scientific american}, 267\penalty0 (1):\penalty0 66--73, 1992.

\bibitem[Hu et~al.(2018)Hu, Shen, and Sun]{hu2018squeeze}
Jie Hu, Li~Shen, and Gang Sun.
\newblock Squeeze-and-excitation networks.
\newblock In \emph{Proceedings of the IEEE conference on computer vision and
  pattern recognition}, pp.\  7132--7141, 2018.

\bibitem[Huang et~al.(2017)Huang, Liu, Van Der~Maaten, and
  Weinberger]{huang2017densely}
Gao Huang, Zhuang Liu, Laurens Van Der~Maaten, and Kilian~Q Weinberger.
\newblock Densely connected convolutional networks.
\newblock In \emph{Proceedings of the IEEE conference on computer vision and
  pattern recognition}, pp.\  4700--4708, 2017.

\bibitem[Huizinga \& Clune(2018)Huizinga and Clune]{huizinga2018evolving}
Joost Huizinga and Jeff Clune.
\newblock Evolving multimodal robot behavior via many stepping stones with the
  combinatorial multi-objective evolutionary algorithm.
\newblock \emph{arXiv preprint arXiv:1807.03392}, 2018.

\bibitem[Ioffe \& Szegedy(2015)Ioffe and Szegedy]{ioffe2015batch}
Sergey Ioffe and Christian Szegedy.
\newblock Batch normalization: Accelerating deep network training by reducing
  internal covariate shift.
\newblock In \emph{International conference on machine learning}, pp.\
  448--456. PMLR, 2015.

\bibitem[Jaderberg et~al.(2017)Jaderberg, Dalibard, Osindero, Czarnecki,
  Donahue, Razavi, Vinyals, Green, Dunning, Simonyan,
  et~al.]{jaderberg2017population}
Max Jaderberg, Valentin Dalibard, Simon Osindero, Wojciech~M Czarnecki, Jeff
  Donahue, Ali Razavi, Oriol Vinyals, Tim Green, Iain Dunning, Karen Simonyan,
  et~al.
\newblock Population based training of neural networks.
\newblock \emph{arXiv preprint arXiv:1711.09846}, 2017.

\bibitem[Krawiec \& Liskowski(2015)Krawiec and Liskowski]{krawiec2015automatic}
Krzysztof Krawiec and Pawe{\l} Liskowski.
\newblock Automatic derivation of search objectives for test-based genetic
  programming.
\newblock In \emph{European Conference on Genetic Programming}, pp.\  53--65.
  Springer, 2015.

\bibitem[Krizhevsky et~al.(2009)Krizhevsky, Hinton,
  et~al.]{krizhevsky2009learning}
Alex Krizhevsky, Geoffrey Hinton, et~al.
\newblock Learning multiple layers of features from tiny images.
\newblock \emph{Tech. Report}, 2009.

\bibitem[La~Cava \& Moore(2018)La~Cava and Moore]{la2018behavioral}
William La~Cava and Jason Moore.
\newblock Behavioral search drivers and the role of elitism in soft robotics.
\newblock In \emph{ALIFE 2018: The 2018 Conference on Artificial Life}, pp.\
  206--213. MIT Press, 2018.

\bibitem[La~Cava \& Moore(2020{\natexlab{a}})La~Cava and Moore]{la2020genetic}
William La~Cava and Jason~H Moore.
\newblock Genetic programming approaches to learning fair classifiers.
\newblock In \emph{Proceedings of the 2020 Genetic and Evolutionary Computation
  Conference}, pp.\  967--975, 2020{\natexlab{a}}.

\bibitem[La~Cava \& Moore(2020{\natexlab{b}})La~Cava and Moore]{la2020learning}
William La~Cava and Jason~H Moore.
\newblock Learning feature spaces for regression with genetic programming.
\newblock \emph{Genetic Programming and Evolvable Machines}, 21\penalty0
  (3):\penalty0 433--467, 2020{\natexlab{b}}.

\bibitem[La~Cava et~al.(2016)La~Cava, Spector, and Danai]{la2016epsilon}
William La~Cava, Lee Spector, and Kourosh Danai.
\newblock Epsilon-lexicase selection for regression.
\newblock In \emph{Proceedings of the Genetic and Evolutionary Computation
  Conference 2016}, pp.\  741--748, 2016.

\bibitem[La~Cava et~al.(2019)La~Cava, Helmuth, Spector, and
  Moore]{la2019probabilistic}
William La~Cava, Thomas Helmuth, Lee Spector, and Jason~H Moore.
\newblock A probabilistic and multi-objective analysis of lexicase selection
  and $\varepsilon$-lexicase selection.
\newblock \emph{Evolutionary Computation}, 27\penalty0 (3):\penalty0 377--402,
  2019.

\bibitem[LeCun et~al.(2015)LeCun, Bengio, and Hinton]{lecun2015deep}
Yann LeCun, Yoshua Bengio, and Geoffrey Hinton.
\newblock Deep learning.
\newblock \emph{nature}, 521\penalty0 (7553):\penalty0 436--444, 2015.

\bibitem[Liskowski et~al.(2015)Liskowski, Krawiec, Helmuth, and
  Spector]{liskowski2015comparison}
Pawel Liskowski, Krzysztof Krawiec, Thomas Helmuth, and Lee Spector.
\newblock Comparison of semantic-aware selection methods in genetic
  programming.
\newblock In \emph{Proceedings of the Companion Publication of the 2015 Annual
  Conference on Genetic and Evolutionary Computation}, pp.\  1301--1307, 2015.

\bibitem[Liu et~al.(2017)Liu, Simonyan, Vinyals, Fernando, and
  Kavukcuoglu]{liu2017hierarchical}
Hanxiao Liu, Karen Simonyan, Oriol Vinyals, Chrisantha Fernando, and Koray
  Kavukcuoglu.
\newblock Hierarchical representations for efficient architecture search.
\newblock \emph{arXiv preprint arXiv:1711.00436}, 2017.

\bibitem[Liu et~al.(2020)Liu, Gao, and Yin]{liu2020improved}
Yanli Liu, Yuan Gao, and Wotao Yin.
\newblock An improved analysis of stochastic gradient descent with momentum.
\newblock \emph{arXiv preprint arXiv:2007.07989}, 2020.

\bibitem[Loshchilov \& Hutter(2017)Loshchilov and Hutter]{loshchilov2017sgdr}
Ilya Loshchilov and Frank Hutter.
\newblock {SGDR:} stochastic gradient descent with warm restarts.
\newblock In \emph{5th International Conference on Learning Representations
  (ICLR)}, 2017.

\bibitem[Luo et~al.(2018)Luo, Xiong, Liu, and Sun]{luo2018adaptive}
Liangchen Luo, Yuanhao Xiong, Yan Liu, and Xu~Sun.
\newblock Adaptive gradient methods with dynamic bound of learning rate.
\newblock In \emph{International Conference on Learning Representations}, 2018.

\bibitem[Metevier et~al.(2019)Metevier, Saini, and
  Spector]{metevier2019lexicase}
Blossom Metevier, Anil~Kumar Saini, and Lee Spector.
\newblock Lexicase selection beyond genetic programming.
\newblock In \emph{Genetic Programming Theory and Practice XVI}, pp.\
  123--136. Springer, 2019.

\bibitem[Miikkulainen et~al.(2019)Miikkulainen, Liang, Meyerson, Rawal, Fink,
  Francon, Raju, Shahrzad, Navruzyan, Duffy, et~al.]{miikkulainen2019evolving}
Risto Miikkulainen, Jason Liang, Elliot Meyerson, Aditya Rawal, Daniel Fink,
  Olivier Francon, Bala Raju, Hormoz Shahrzad, Arshak Navruzyan, Nigel Duffy,
  et~al.
\newblock Evolving deep neural networks.
\newblock In \emph{Artificial intelligence in the age of neural networks and
  brain computing}, pp.\  293--312. Elsevier, 2019.

\bibitem[Miller et~al.(1995)Miller, Goldberg, et~al.]{miller1995genetic}
Brad~L Miller, David~E Goldberg, et~al.
\newblock Genetic algorithms, tournament selection, and the effects of noise.
\newblock \emph{Complex systems}, 9\penalty0 (3):\penalty0 193--212, 1995.

\bibitem[Netzer et~al.(2011)Netzer, Wang, Coates, Bissacco, Wu, and
  Ng]{netzer2011reading}
Yuval Netzer, Tao Wang, Adam Coates, Alessandro Bissacco, Bo~Wu, and Andrew~Y
  Ng.
\newblock Reading digits in natural images with unsupervised feature learning.
\newblock \emph{NeurIPS Workshop on Deep Learning and Unsupervised Feature
  Learning}, 2011.

\bibitem[Pawe{\l}czyk et~al.(2018)Pawe{\l}czyk, Kawulok, and
  Nalepa]{pawelczyk2018genetically}
Krzysztof Pawe{\l}czyk, Michal Kawulok, and Jakub Nalepa.
\newblock Genetically-trained deep neural networks.
\newblock In \emph{Proceedings of the Genetic and Evolutionary Computation
  Conference Companion}, pp.\  63--64, 2018.

\bibitem[Real et~al.(2017)Real, Moore, Selle, Saxena, Suematsu, Tan, Le, and
  Kurakin]{real2017large}
Esteban Real, Sherry Moore, Andrew Selle, Saurabh Saxena, Yutaka~Leon Suematsu,
  Jie Tan, Quoc~V Le, and Alexey Kurakin.
\newblock Large-scale evolution of image classifiers.
\newblock In \emph{International Conference on Machine Learning}, pp.\
  2902--2911. PMLR, 2017.

\bibitem[Real et~al.(2019)Real, Aggarwal, Huang, and Le]{real2019regularized}
Esteban Real, Alok Aggarwal, Yanping Huang, and Quoc~V Le.
\newblock Regularized evolution for image classifier architecture search.
\newblock In \emph{Proceedings of the aaai conference on artificial
  intelligence}, volume~33, pp.\  4780--4789, 2019.

\bibitem[Rumelhart et~al.(1985)Rumelhart, Hinton, and
  Williams]{rumelhart1985learning}
David~E Rumelhart, Geoffrey~E Hinton, and Ronald~J Williams.
\newblock Learning internal representations by error propagation.
\newblock Technical report, California Univ San Diego La Jolla Inst for
  Cognitive Science, 1985.

\bibitem[Russakovsky et~al.(2015)Russakovsky, Deng, Su, Krause, Satheesh, Ma,
  Huang, Karpathy, Khosla, Bernstein, et~al.]{russakovsky2015imagenet}
Olga Russakovsky, Jia Deng, Hao Su, Jonathan Krause, Sanjeev Satheesh, Sean Ma,
  Zhiheng Huang, Andrej Karpathy, Aditya Khosla, Michael Bernstein, et~al.
\newblock Imagenet large scale visual recognition challenge.
\newblock \emph{International journal of computer vision}, 115\penalty0
  (3):\penalty0 211--252, 2015.

\bibitem[Sandler et~al.(2018)Sandler, Howard, Zhu, Zhmoginov, and
  Chen]{sandler2018mobilenetv2}
Mark Sandler, Andrew Howard, Menglong Zhu, Andrey Zhmoginov, and Liang-Chieh
  Chen.
\newblock Mobilenetv2: Inverted residuals and linear bottlenecks.
\newblock In \emph{Proceedings of the IEEE conference on computer vision and
  pattern recognition}, pp.\  4510--4520, 2018.

\bibitem[Simonyan \& Zisserman(2015)Simonyan and Zisserman]{simonyan15very}
Karen Simonyan and Andrew Zisserman.
\newblock Very deep convolutional networks for large-scale image recognition.
\newblock In \emph{International Conference on Learning Representations}, 2015.

\bibitem[Spector(2012)]{spector2012assessment}
Lee Spector.
\newblock Assessment of problem modality by differential performance of
  lexicase selection in genetic programming: a preliminary report.
\newblock In \emph{Proceedings of the 14th annual conference companion on
  Genetic and evolutionary computation}, pp.\  401--408, 2012.

\bibitem[Stanley \& Miikkulainen(2002)Stanley and
  Miikkulainen]{stanley2002evolving}
Kenneth~O Stanley and Risto Miikkulainen.
\newblock Evolving neural networks through augmenting topologies.
\newblock \emph{Evolutionary computation}, 10\penalty0 (2):\penalty0 99--127,
  2002.

\bibitem[Such et~al.(2017)Such, Madhavan, Conti, Lehman, Stanley, and
  Clune]{such2017deep}
Felipe~Petroski Such, Vashisht Madhavan, Edoardo Conti, Joel Lehman, Kenneth~O
  Stanley, and Jeff Clune.
\newblock Deep neuroevolution: Genetic algorithms are a competitive alternative
  for training deep neural networks for reinforcement learning.
\newblock \emph{arXiv preprint arXiv:1712.06567}, 2017.

\bibitem[Sutskever et~al.(2013)Sutskever, Martens, Dahl, and
  Hinton]{sutskever2013importance}
Ilya Sutskever, James Martens, George Dahl, and Geoffrey Hinton.
\newblock On the importance of initialization and momentum in deep learning.
\newblock In \emph{International conference on machine learning}, pp.\
  1139--1147. PMLR, 2013.

\bibitem[Tan \& Le(2019)Tan and Le]{tan2019efficientnet}
Mingxing Tan and Quoc Le.
\newblock Efficientnet: Rethinking model scaling for convolutional neural
  networks.
\newblock In \emph{International Conference on Machine Learning}, pp.\
  6105--6114. PMLR, 2019.

\bibitem[Wilson et~al.(2017)Wilson, Roelofs, Stern, Srebro, and
  Recht]{wilson2017marginal}
Ashia~C Wilson, Rebecca Roelofs, Mitchell Stern, Nathan Srebro, and Benjamin
  Recht.
\newblock The marginal value of adaptive gradient methods in machine learning.
\newblock In \emph{Proceedings of the 31st International Conference on Neural
  Information Processing Systems}, pp.\  4151--4161, 2017.

\bibitem[Wu \& He(2018)Wu and He]{wu2018group}
Yuxin Wu and Kaiming He.
\newblock Group normalization.
\newblock In \emph{Proceedings of the European conference on computer vision
  (ECCV)}, pp.\  3--19, 2018.

\bibitem[Xie \& Yuille(2017)Xie and Yuille]{xie2017genetic}
Lingxi Xie and Alan Yuille.
\newblock Genetic cnn.
\newblock In \emph{Proceedings of the IEEE international conference on computer
  vision}, pp.\  1379--1388, 2017.

\bibitem[Zhang et~al.(2017)Zhang, Bengio, Hardt, Recht, and
  Vinyals]{zhang2017understanding}
Chiyuan Zhang, Samy Bengio, Moritz Hardt, Benjamin Recht, and Oriol Vinyals.
\newblock Understanding deep learning requires rethinking generalization.
\newblock In \emph{5th International Conference on Learning Representations
  (ICLR)}, 2017.

\end{thebibliography}
\bibliographystyle{iclr2022_conference}

\newpage
\appendix

\section{Lexicase Selection} \label{sec:lexicase}

Lexicase selection is a parent selection method in population-based stochastic
search algorithms such as genetic
programming~\citep{helmuth2014solving,spector2012assessment}. The lexicase
selection algorithm is outlined in Alg.~\ref{alg:lexi}.

\begin{figure}[h]
    \centering
    \begin{minipage}{.9\linewidth}
        \begin{algorithm}[H]
            \KwData{
                \begin{itemize}
                    \item \texttt{cases} - randomly shuffled sequence of data samples to be used in selection
                    \item \texttt{candidates} - the entire population of programs
                \end{itemize}
            }
            \KwResult{
                \begin{itemize}
                    \item an individual program to be used as a parent
                \end{itemize}
            }

            \For{\texttt{case} in \texttt{cases}}{
                \texttt{candidates} $\gets$ the subset of the current \texttt{candidates} that have exactly best performance on \texttt{case}

                \If{\texttt{candidates} contains only one single \texttt{candidate}}{\KwRet{\texttt{candidate}}}
            }

            \KwRet{a randomly selected \texttt{candidate} in \texttt{candidates}}

            \caption{Lexicase selection to select one parent program in genetic programming}
            \label{alg:lexi}
        \end{algorithm}
    \end{minipage}
\end{figure}

The key idea of lexicase selection is that each selection event considers a
randomly shuffled sequence of training cases. With the specific ordering, only
individuals whose error is minimal among all the already-considered cases are
allowed to survive. Moving forward through the sequence of cases, the selection
is done when there is only one candidate left or until all the training cases
have been gone through, in which case we select randomly from the remaining
candidates.

Since the ordering of training cases is randomized for each selection event,
every training case gets the opportunity to be prioritized when being put at the
beginning of the sequence. As a result, lexicase selection sometimes selects
specialist individuals that perform poorly on average but perform better than
many individuals on one or more other cases.

\section{Implementation Details}\label{sec:imp}

Each network architecture with baseline SGD training and its corresponding
counterpart with gradient lexicase selection are trained with identical
experimental schemes. We use SGD with momentum instead of the popular adaptive
methods (such as Adam) because, despite the popularity of those methods, some
recent works~\citep{luo2018adaptive,wilson2017marginal} observe that the
solutions found by those methods actually generalize worse (often significantly
worse) than SGD. We did experiments with Adam and tuned the learning rate for
several trials, but the results are significantly worse than the SGD
counterpart. This work focuses more on the generalization performance rather
than the training speed, so we follow the common practice to use SGD with
momentum for both baseline training and SubGD. However, it is very likely that
some most recent optimization methods, such as \citet{luo2018adaptive}, can
achieve faster training as well as the same generalization performance as SGD.

We follow standard practices and perform data augmentation with random cropping
with padding and perform random horizontal flipping during the training phase
(no augmentation is used during selection phase). The input images are
normalized through mean RGB-channel subtraction for all the phases. For both
baseline and lexicase, we use SGD with momentum of $0.9$. For lexicase, we use
the Reset Momentum option that re-initialize the momentum parameters for each
epoch, which is explained in detail later in Sec.~\ref{sec:mom}.

The batch size is set to $128$ for CIFAR-10 and $64$ for CIFAR-100 and SVHN. The
initial learning rate is set to $0.1$ and tuned by using Cosine
Annealing~\citep{loshchilov2017sgdr}.

We set the total number of epochs as $200$ for baseline training and as
$200(p+1)$ for gradient lexicase selection, where $p$ is the size of population.
For each epoch in lexicase, the mutation only uses $1/p$ of the training data to
train each model instance, so we keep the total iterations of weight update of
lexicase training similar to baseline training to ensure convergence.

\end{document}